\documentclass[letterpaper, 10 pt, conference]{ieeeconf}  

\IEEEoverridecommandlockouts                              

\usepackage{amsfonts}
\usepackage{amsmath,amssymb,amsfonts}
\usepackage{url}
\usepackage{etoolbox}
\usepackage{graphicx}
\usepackage{optidef}
\usepackage{algpseudocode}
\usepackage{mathtools,amssymb}
\usepackage{svg}
\usepackage{ulem}
\usepackage{stix}
\usepackage{amsmath,mleftright,mathtools}
\DeclareMathAlphabet\mathbfcal{OMS}{cmsy}{b}{n}
\usepackage{hyperref}
\usepackage{stfloats}
\usepackage{lipsum}
\usepackage{comment}
\usepackage{subcaption} 
\usepackage[linesnumbered,ruled,vlined,commentsnumbered]{algorithm2e}
\usepackage{svg}

\usepackage{colortbl}
\usepackage{siunitx}

\usepackage{placeins}

\title{\LARGE \bf
Autonomous Reactive Masonry Construction using Collaborative Heterogeneous Aerial Robots with Experimental Demonstration
}

\author{Marios-Nektarios Stamatopoulos, Elias Small, Shridhar Velhal, Avijit Banerjee and George Nikolakopoulos
}

\begin{document}

\maketitle

\thispagestyle{empty}
\pagestyle{empty}

\begin{abstract}
%
This article presents a fully autonomous aerial masonry construction framework using heterogeneous unmanned aerial vehicles (UAVs), supported by experimental validation. Two specialized UAVs were developed for the task: (i) a brick-carrier UAV equipped with a ball-joint actuation mechanism for precise brick manipulation, and (ii) an adhesion UAV integrating a servo-controlled valve and extruder nozzle for accurate adhesion application. The proposed framework employs a reactive mission planning unit that combines a dependency graph of the construction layout with a conflict graph to manage simultaneous task execution, while hierarchical state machines ensure robust operation and safe transitions during task execution. Dynamic task allocation allows real-time adaptation to environmental feedback, while minimum-jerk trajectory generation ensures smooth and precise UAV motion during brick pickup and placement. Additionally, the brick-carrier UAV employs an onboard vision system that estimates brick poses in real time using ArUco markers and a least-squares optimization filter, enabling accurate alignment during construction. To the best of the authors’ knowledge, this work represents the first experimental demonstration of fully autonomous aerial masonry construction using heterogeneous UAVs, where one UAV precisely places the bricks while another autonomously applies adhesion material between them. The experimental results supported by the video showcase the effectiveness of the proposed framework and demonstrate its potential to serve as a foundation for future developments in autonomous aerial robotic construction. A video with the framework and the experimental mission is available at \href{https://youtu.be/OBh6SibQh9U}{https://youtu.be/OBh6SibQh9U}.
\end{abstract}

\section{Introduction}

Recent progress in robotic autonomy, supported by advanced sensing technologies and planning frameworks, is reshaping the construction sector \cite{advancementsRoboticsConstr}. Robotics is increasingly recognized as a disruptive enabler, with the potential to boost productivity, improve workplace safety \cite{saidi2016}, and address the challenges of autonomous coordination in high-risk environments while alleviating human workload.  

Research efforts have investigated a variety of robotic construction platforms, including gantry-based systems that leverage both additive manufacturing techniques \cite{largeScale3DPrint} and conventional bricklaying methods \cite{BRIX}. Other approaches have employed ground-based autonomous systems or mobile robots with manipulators, which can contribute to construction workflows by operating on elevated rails \cite{brickLabyrinth} or performing cement extrusion during motion \cite{printingWhileMoving}.  

Despite these promising developments, such terrestrial systems face significant constraints, particularly limited reach in critical or infrastructure-deficient sites, along with demanding deployment logistics. An attractive alternative lies in the use of Unmanned Aerial Vehicles (UAVs) as autonomous builders. Extending additive manufacturing concepts, UAVs can serve as airborne robotic fabricators by extruding material in flight \cite{nature_aerial_AM,stamatopoulos2025ExperimentAutCon,stamatopoulos2024collaborative}. Alternatively, they can execute autonomous pick-and-place operations to assemble large-scale brick structures \cite{flightAssembledArch,Xie2025aerial}.  
While initial work on UAV-based autonomous bricklaying has established feasibility \cite{feasibility}, prior studies have largely focused on brick pickup and placement and on the design of aerial manipulation mechanisms. Yet, for practical full-scale autonomous construction, brick stacking alone is insufficient: ensuring structural integrity necessitates the integration of bonding processes, such as the application of cementitious materials or mortar for reinforcement and adhesion.

\begin{figure}[t]
    \centering
    \includegraphics[width=\linewidth]{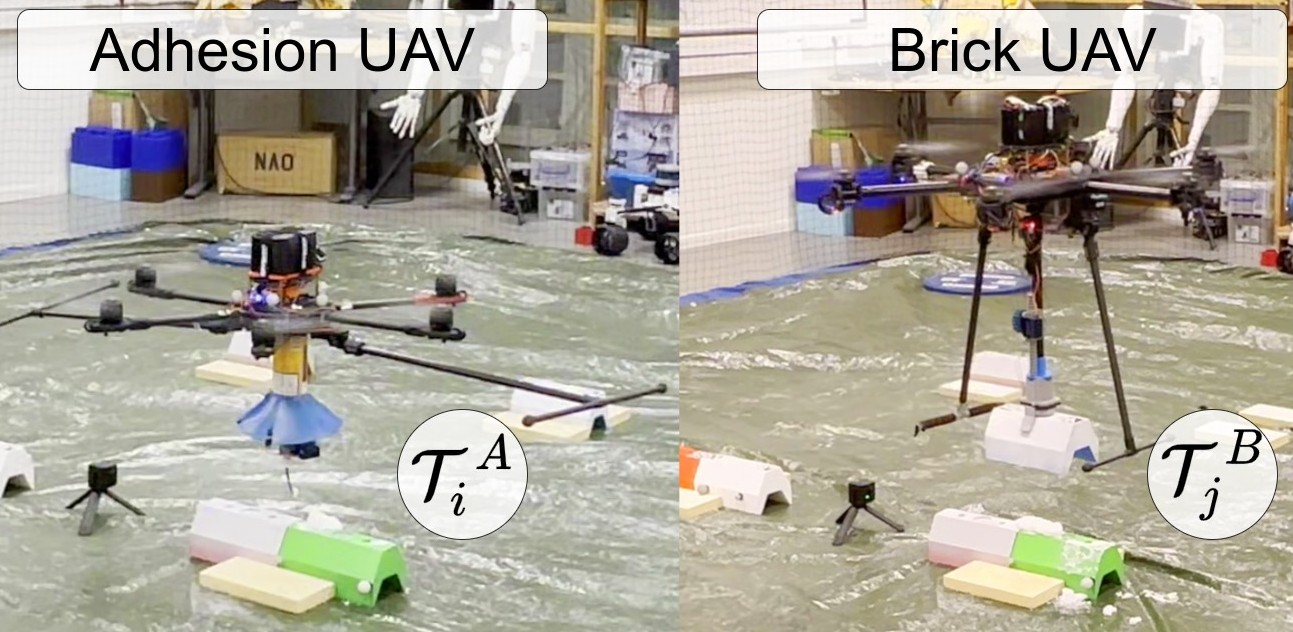}
    \caption{Concept figure of the proposed framework where the brick UAV $\mathcal{U}^B$ is placing the brick task $\mathcal{T}^B_i$ (left) and the adhesion robot $\mathcal{U}^A$ is executing the adhesion task $\mathcal{T}^A_j$.}
    \label{fig:concept}
    \vspace{-6mm}
\end{figure}

\subsection{Related Work }

The cooperative wall construction using multiple UAVs was initially presented in \cite{flightAssembledArch}, which relied on a construction blueprint with the exact positioning of each brick. A centralized reactive assignment approach had been used to assign tasks to the available UAVs to avoid inter-UAV collisions.  However, the system required manual intervention as adhesion mortar was placed manually before pickup.

In the MBZIRC 2020 competition, the second challenge focused on cooperative construction, requiring collaborative efforts from ground and aerial robots. In \cite{mbzircWallSaska}, each UAV is assigned specific segments of the wall to construct, along with corresponding stacks of bricks to retrieve from. This division of tasks optimizes UAV movements and reduces the likelihood of overlap or collision during the building process. In \cite{mbzircMetaheuristicWallplanner}, the construction problem is formulated as a variant of the Team Orienteering Problem, incorporating precedence and concurrency constraints to ensure safe and orderly brick placement among UAVs. Near-optimal building plans are then generated using a meta-heuristic approach, the Greedy Randomized Adaptive Search Procedure (GRASP). Similarly, \cite{mbzircBarbara} introduces a hierarchical task planning and coordination framework that enables effective task allocation and scheduling across different brick types. This decentralized strategy demonstrates cooperation between a mobile robot and up to three UAVs, optimizing the utilization of each robot’s capabilities. Recently, in \cite{Xie2025Brickpilot}, an autonomous system for autonomously recognising, grasping, and placing real-scale bricks is demonstrated.  
The aforementioned literature does not consider the problem of placing mortar autonomously. The brick placement and mortar placement are heterogeneous tasks and demand different kinds of UAVs. The heterogeneous system demands separate attention.  
The various challenges and methodologies for masonry construction had been discussed in \cite{elkhapery2022exploratory}. 
Recently, safe and cooperative planning for masonry construction has been proposed in \cite{stamatopoulos2025safety}, but it lacks real-world experiments. 

\subsection{Contributions}
The primary contribution of this work stems from the design and experimental validation of a fully autonomous aerial masonry construction framework. To this end, two heterogeneous UAVs were specifically developed and deployed: (i) a brick-carrier UAV, responsible for brick pickup and placement, equipped with a vision system, ArUco marker–based brick detection, and a ball-joint actuation mechanism for reliable aerial manipulation; and (ii) an adhesion UAV, featuring a servo-controlled valve and extruder nozzle for precise adhesion application. 

Towards achieving this goal, the article deploys a reactive planning framework that advances autonomy in multi-robot construction. A mission planner leverages a dependency graph of the construction layout together with a conflict graph to manage simultaneous task execution. Tasks are dynamically allocated to the UAVs according to situational context and real-time environmental feedback, ensuring robustness in collaborative operations. To ensure robust operation in uncertain conditions, execution is governed by hierarchical state machines that monitor system states, handle contingencies, and enforce safe transitions between sub-tasks. Smooth and safe flight is further achieved through minimum-jerk trajectory generation, which enables the carrier UAV to establish the pickup-drop operation with high terminal accuracy, while maintaining ground clearance.

Another contribution lies in accurate and reliable brick manipulation enabled by robust perception. The pose of each brick is estimated in real time using onboard vision, supported by ArUco markers, and a nonlinear least-squares optimization filter, providing stable estimates of both position and orientation. This integration of sensing, filtering, and state-machine–driven execution ensures precise alignment and reliable task completion. To the best of the authors’ knowledge, this work constitutes the first experimental demonstration of fully autonomous aerial masonry construction employing heterogeneous robotic systems for coordinated brick placement and adhesion application. The experimental results are presented, and a demonstration of the proposed framework is provided in the accompanying video.

\section{System Architecture}\label{sec:systemArch}
\begin{figure}[bhp!]
    \centering
    \includegraphics[width=0.99\linewidth]{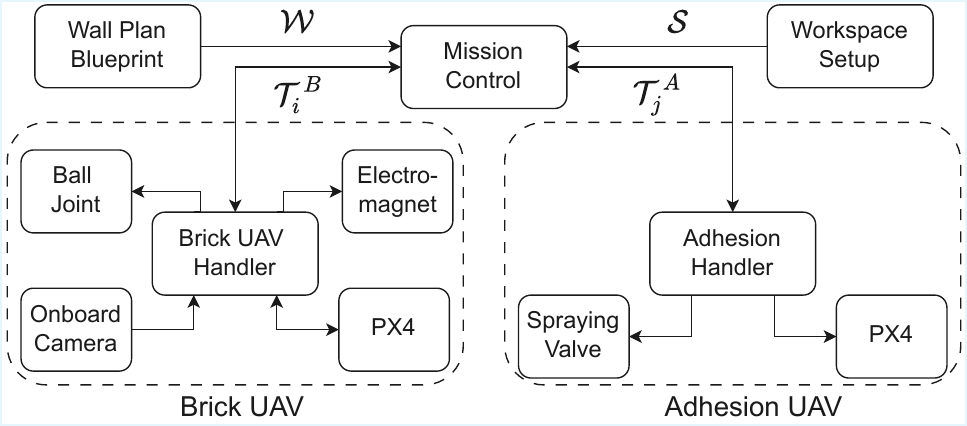}
    \caption{Overall Block Diagram of the proposed framework.}
    \label{fig:blockDiagram}
    \vspace{-5mm}
\end{figure}
A schematic representation of the overall architecture for the proposed autonomy framework in a block diagram format is shown in Fig. \ref{fig:blockDiagram}. Initially, the wall plan blueprint $\mathcal{W}$ containing all the brick and adhesion placement tasks for the UAVs is given to the mission control in addition to the setup $\mathcal{S}$ of the current workspace, including the approximate pickup locations of each brick. Then, the mission control identifies all dependencies and conflicts between the tasks and generates both the dependency graph $\mathcal{G}^d$ and the conflict graph $\mathcal{G}^c$. The brick $\mathcal{T}^B_i$ and adhesion $\mathcal{T}^A_j$ tasks are assigned to the UAVs in a centralized manner by the mission control module, which is executed on a third module. 

The system consists of two types of UAVs, the brick UAV $\mathcal{U}^B$, where each whole motion and actuation is operated on the handler module, which communicates with the mission controller and receives the assigned task. It also gets the brick position estimation from the onboard camera and actuates both the ball joint locking mechanism and the electromagnet at the tip in order to pick up the brick. Finally, it interfaces with the onboard flight controller, which is responsible for the motion of the UAV.
Similarly, the adhesion UAV $\mathcal{U}^A$ consists of the main handler, which receives the adhesion task from the mission controller and coordinates the execution of the adhesion. By actuating the onboard servo attached to the spraying valve, it controls the spraying of the adhesion material.

\section{Mission Planner and Task Assignment}
\subsubsection{Dependency Graph $\mathcal{G}^d$}
After the wall plan blueprint $\mathcal{W}$ is given to the mission control, a systematic process is carried out in order to calculate the dependency graph $\mathcal{G}^d$ in which brick pairs are examined to identify structural dependencies. A dependency from brick $B_i$ to brick $B_j$ exists if $B_i$ is positioned directly above $B_j$.
For every such dependent pair, the horizontal overlap between the two bricks is computed. Using this overlap, an adhesion task $A_m$ is defined, starting at the overlap’s initial position and extending across its width.
Subsequently, the dependency graph $\mathcal{G}^d$ is augmented with two directed edges: $(B_i, A_m)$ and $(A_m, B_j)$, which capture the structural connection between the upper and lower bricks mediated by the adhesion task.

\subsubsection{Conflict Graph $\mathcal{G}^c$}
On top of this, another step is carried out to ensure the safety between the UAVs. Given the operational workspace of each robot, there exists a risk of two robots being in proximity, potentially causing a collision during simultaneous task execution. To mitigate this risk, a minimum clearance $r_c \in \mathbb{R}$ between UAVs is enforced throughout the mission. This requirement is integrated into the planning stage by evaluating the distances between all task locations and applying concurrency constraints to tasks that do not satisfy the clearance condition. The resulting conflicts are categorized into sets, with each set grouping tasks that are mutually conflicting. Specifically, two sets are defined, $\mathcal{C}^B_B$ and $\mathcal{C}^A_B$, as follows:
\begin{align}
\mathcal{C}^B_B &= { (B_i, B_j) : \text{dist}(B_i, B_j)\leq r_c} \\
\mathcal{C}^A_B &= { (A_i, B_j) : \text{dist}(A_i, B_j)\leq r_c}
\end{align}
After constructing these sets, all pairs that exhibit any form of task dependency are removed, thereby reducing the number of enforced constraints.

\subsubsection{Reactive Assignment}
Adhesion and brick tasks are allocated to the robots in a reactive manner by the mission planner, enabling the system to adapt seamlessly to delays, disturbances, and potential failures encountered under real-world operating conditions. The availability of tasks is governed by the dependency graph $\mathcal{G}^d$, which is continuously updated as each task is completed, thereby ensuring that the mission plan evolves dynamically in response to execution progress.
A task $\mathcal{T}^A_i$ or $\mathcal{T}^B_j$ is considered available if it has no descendants in $\mathcal{G}^d$ or if all of its descendant tasks have already been completed. 
For safety, a task may only be assigned if it does not conflict with any task currently in progress. Accordingly, whenever a new task is assigned, the conflict graph $\mathcal{G}^c$ is updated to mark all conflicting tasks as unavailable.
An overview of the assignment process is shown in Algorithm~\ref{alg:taskAssign}.

\SetKwIF{If}{ElseIf}{Else}{if}{then}{else if}{else}{end if}
{\normalem
\begin{algorithm}[tbp!]
    \caption{Reactive Task Assignment}
    \label{alg:taskAssign}
    \small
    \DontPrintSemicolon
    \LinesNumbered
    \SetAlgoNlRelativeSize{0}
    \SetNlSty{textbf}{}{:}
    
    \KwIn{Dependency Graph $\mathcal{G}^d$ and Conflict Graph $\mathcal{G}^c$}
    \KwOut{Assigned tasks $\mathcal{T}^A_i, \mathcal{T}^B_j$}
    
    \While{mission is not completed}{
        $\mathcal{T}^A_{av}, \mathcal{T}^B_{av} \gets$ Available tasks as per dependency graph $\mathcal{G}^d$\;
        $\mathcal{T}^A_{av}, \mathcal{T}^B_{av} \gets$ Prune conflicting tasks\;
        $\mathcal{U}^A_{av}, \mathcal{U}^B_{av} \gets$ Idle Adhesion and Brick UAVs\;
        
        \If{$|\mathcal{U}^B_{av}| > 0$ and $|\mathcal{T}^B_{av}| > 0$}{
            Assign pickup tasks $\mathcal{T}^B_{av}$ to available $\mathcal{U}^B_{av}$ UAVs\;
        }
        
        \If{$|\mathcal{U}^A_{av}| > 0$ and $|\mathcal{T}^A_{av}| > 0$}{
            Assign adhesion tasks $\mathcal{T}^A_{av}$ to available $\mathcal{U}^A_{av}$ UAVs\;
        }
        
        Dynamically update the conflict graph $\mathcal{G}^c$\;
    }
\end{algorithm}
}

\section{Pickup and Drop-off UAV}

\subsection{Platform Description}
\begin{figure}
    \centering
    \includegraphics[width=0.9\linewidth]{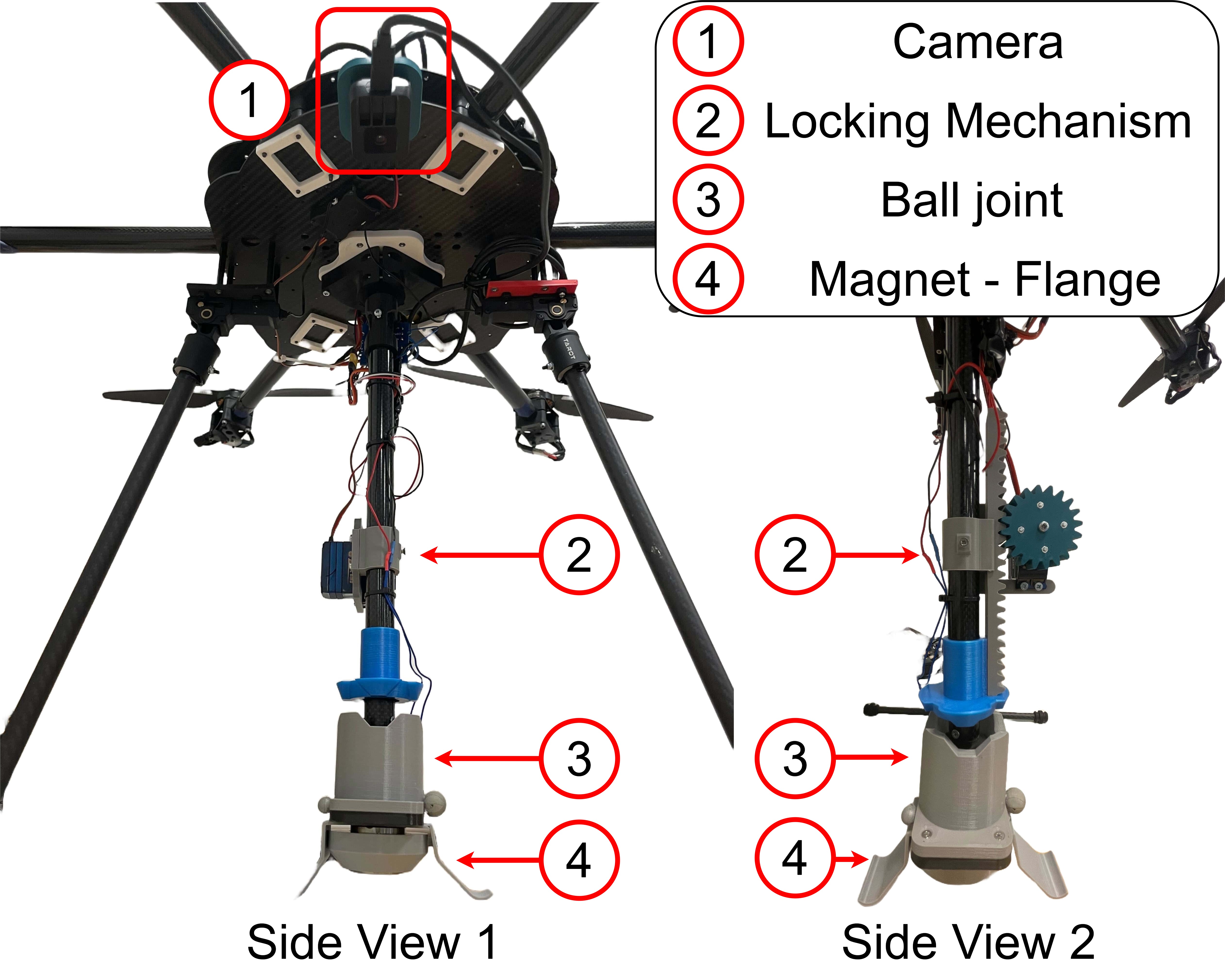}
    \caption{Brick UAV $\mathcal{U}^B$ Pickup Mechanism }
    \label{fig:placeholder}
    \vspace{-5mm}
\end{figure}

The UAV platform employed for brick pickup and placement is a custom-designed and assembled hexacopter with retractable landing gear and has a wingspan of 0.8 m and a total mass of 4.5 kg, including batteries. A rigidly mounted rod is attached to the underside of the UAV, carrying an electromagnet at its tip for grasping bricks equipped with a central metallic plate. The electromagnet is connected via a ball joint, which can be mechanically locked using a 3D-printed slider actuated by a servo. During flight, the ball joint remains locked to ensure stability; however, it is unlocked when the UAV approaches a brick for pickup or during the placement of a carried brick. This design decouples the orientation of the electromagnet from the UAV’s motion in situations requiring precision and delicacy, reducing sudden disturbances that could otherwise make pickup and drop-off unreliable. Furthermore, allowing the ball joint to move freely during these operations introduces a small degree of compliance, enabling the electromagnet to self-align with the metallic plate even in cases of slight misalignment, thereby facilitating smoother engagement with the brick’s reception area. \textcolor{black}{To further aid the consistency of contact between the electromagnet and the brick, a conical ring was attached, as well as an angled flange around the electromagnet to guide it in place during pick-up. The flange also prevents the unexpected rotation of the brick during flight. A downward-facing camera is attached to the body of the UAV for brick position estimation. Fig. \ref{fig:placeholder} shows the relative position of the camera and the construction of the pick-up mechanism.}

\subsection{Sequence}
The autonomous aerial manipulation process follows a structured sequence that integrates both approximate task knowledge and vision-based refinement. At the beginning of each cycle, the UAV remains idle until it is assigned a task $\mathcal{T}^B$ consisting of a pickup brick and a designated drop-off position. Once the assignment is made, the UAV initiates takeoff and ascends to a predefined cruising altitude, which serves as a safe operational layer for subsequent navigation.

Each brick has an approximate pickup location that is known in advance. The UAV navigates laterally until it reaches over the approximate location. Since this estimate is not sufficiently accurate for physical interaction, the UAV employs its onboard camera system and brick estimation algorithm to refine the brick pose. This process involves stabilizing above the approximate site, visually detecting the target brick, and estimating its accurate position and orientation. The refined estimate replaces the initial imprecise location and enables precise alignment of the manipulator with the target. To ensure accurate estimation of the pose, a confidence metric is used for each estimate, where the estimate is considered accurate after a certain threshold $C_{th} \in \mathbb{R}$. In case the confidence is below it or the brick is not visible, the UAV incrementally ascends in order to increase its field of view until the estimation conditions are~satisfied.

Once the pose is confirmed, the UAV aligns directly above the brick and initiates the pickup maneuver. This involves a controlled minimum jerk descent from cruising altitude while the electromagnet is activated. Upon reaching the appropriate pickup height, the UAV engages the end-effector, secures the brick, and subsequently climbs back to cruising altitude to minimize collision risks. A verification stage ensures that the brick has been successfully attached using the onboard camera; in the case that the brick could not be picked up, the estimation and descent process is carried out again.

\begin{figure}
    \centering
    \includegraphics[width=\linewidth]{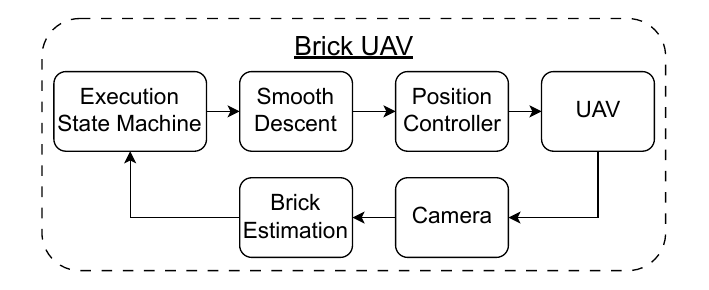}
    \caption{Block diagram of brick UAV $\mathcal{U}^B$ reactive execution.}
    \label{fig:uavBrickBlock}
    \vspace{-5mm}
\end{figure}

After a successful pickup, the UAV proceeds toward the drop-off site. 
%
The drop-off positions are predefined by the wall blueprint, as each adhesion task directly specifies where the corresponding brick is placed. 
These coordinates constitute the construction plan and are directly retrieved by the mission planner. The UAV approaches the assigned drop-off point at cruising altitude. It then descends in a guided manner to the prescribed drop-off height. As soon as the proximity of the end-effector is below a selected threshold $r_{pl}$, it disengages the electromagnet to release the brick and verifies that the brick has been deposited. In case the proximity was not small enough within a given timeout, the UAV hovers again over the drop-off point, and the same sequence is executed. Finally, the UAV ascends back to cruising altitude, from where it either transitions to a new pickup task if one has already been assigned or navigates back to its home position before landing. An overview of the aforementioned architecture for the brick UAV $\mathcal{U}^B$ is shown in Fig. \ref{fig:uavBrickBlock}.

This sequential integration of approximate knowledge, online vision-based refinement, and blueprint-defined placement ensures both flexibility in uncertain pickup environments and accuracy in structured drop-off tasks. It also provides a robust framework for autonomous aerial construction where perception and planning must be tightly coupled.

\subsection{Brick Pose Estimation}

\subsubsection{Brick Setup}
\begin{figure}
    \centering
    \includegraphics[width=0.9\linewidth]{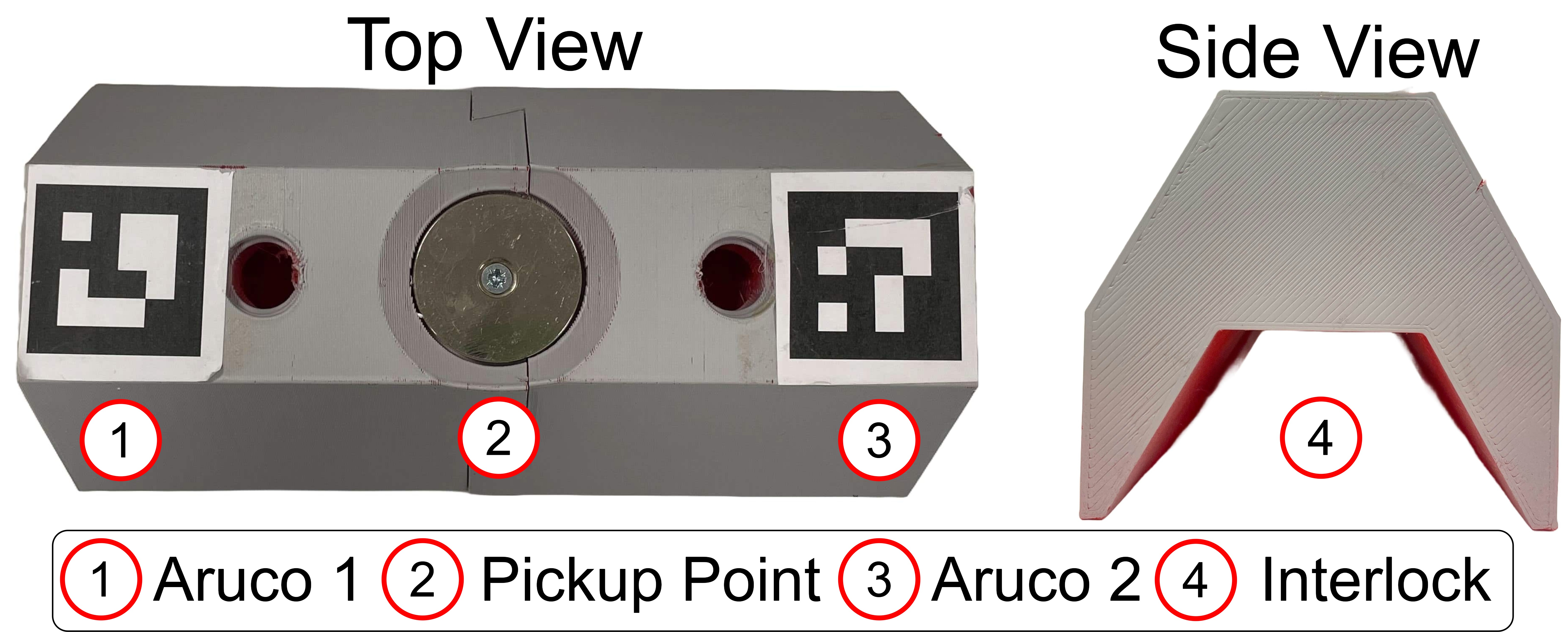}
    \caption{Brick equipped with markers placed on each edge }
    \label{fig:brick}
    \vspace{-5mm}
\end{figure}

Each brick is equipped with a pair of ArUco markers \cite{aruco} that enable identification and tracking through the onboard camera. These markers allow estimation of the brick’s pose, thereby determining its precise pickup location. They are located at the edge of each side of the brick. The identity of each brick is defined by the IDs of the attached ArUco markers. An example is shown in Fig. \ref{fig:brick}, where the brick with ID 0 is represented using the markers with IDs 0 and 1.
These two markers are placed to enhance the efficiency of estimating the brick position from the UAV in cases where one of them is obscured, but also to reinforce accurate estimation of the position of the pickup point, given that the distance and angle between them are known.

\subsubsection{Optimal Filter}
To ensure consistent localization of the brick during manipulation, an optimization scheme was employed to refine the raw camera-based detections of the two Aruco markers placed on its top surface. Although the vision system provides noisy measurements of the marker positions, the physical configuration of the brick introduces strong geometric priors that can be exploited to improve accuracy.
Let $p_1^{\text{meas}}, p_2^{\text{meas}} \in \mathbb{R}^3$ denote the measured 3D positions of the two markers obtained from the onboard camera. Since the markers are rigidly attached to the brick, their true positions $p_1, p_2$ must satisfy two constraints:

\begin{enumerate}
    \item \textbf{Distance constraint:} the markers are mounted with a fixed separation $d$, i.e.,
    \begin{equation}
        \|p_1 - p_2\| = d.
    \end{equation}
    \item \textbf{Planarity constraint:} the brick is placed flat and remains upright during operation. Consequently, the vector $(p_1 - p_2)$ lies in a plane orthogonal to the world $z$-axis, i.e.,
    \begin{equation}
        (p_1 - p_2) \cdot n = 0, \quad n = [0, 0, 1]^\top.
    \end{equation}
\end{enumerate}

Thus, the refinement problem is formulated as a nonlinear least-squares optimization, in which the estimated marker positions $\hat{p}_1, \hat{p}_2$ minimize a weighted sum of three terms: (i) proximity to the raw measurements, (ii) deviation from the fixed inter-marker distance, and (iii) violation of the planarity constraint. The resulting cost function is defined as:
\begin{equation}
\begin{aligned}
    J(\hat{p}_1, \hat{p}_2) = & \;
    \underbrace{w_{\text{prox}} \Big(\|\hat{p}_1 - p_1^{\text{meas}}\|^2 + \|\hat{p}_2 - p_2^{\text{meas}}\|^2\Big)}_{\text{Proximity to measurements}} \\
    & + \underbrace{w_{\text{dist}} \Big(\|\hat{p}_1 - \hat{p}_2\| - d \Big)^2}_{\text{Distance constraint}} \\
    & + \underbrace{w_{\text{plane}} \Big((\hat{p}_1 - \hat{p}_2) \cdot n \Big)^2}_{\text{Planarity constraint}}
\end{aligned}
\end{equation}

where $w_{\text{prox}}, w_{\text{dist}}, w_{\text{plane}}$ are user-defined weights that balance measurement fidelity against constraint satisfaction.

The problem is solved using a quasi-Newton method (BFGS), initialized with the raw detections. The optimization consistently converges to physically plausible marker positions that satisfy the brick’s geometry while remaining close to the measured inputs. If convergence fails, the raw detections are used as a fallback.
This formulation provides a lightweight but effective filtering mechanism that reduces the effect of vision noise while embedding prior geometrical knowledge of the brick.

\subsubsection{Brick Pose Estimation}

Once the individual marker positions are refined, the brick’s pose is estimated using the available tag detections. The algorithm takes as input the measured poses of the two ArUco markers (when available) and outputs the estimated center position and orientation (yaw) of the brick in the camera frame.  
When both tags are detected, their positions are first optimized using the previously described procedure, and the brick center is computed as the midpoint of the two markers. The yaw orientation is then obtained from the vector connecting the markers. If only one tag is visible, a stored geometric offset between the tag and the brick center is applied, and the yaw is extracted from the tag’s orientation with a known correction offset. This design ensures robustness to partial occlusion of the markers while maintaining consistent brick pose estimation. The procedure is summarized in Algorithm~\ref{alg:brickPose}.  

\begin{algorithm}[bhp!]
    \caption{Estimate Brick Pose from Tag Poses}
    \label{alg:brickPose}
    \small
    \DontPrintSemicolon
    \KwIn{Tag poses $T_1$, $T_2$}
    \KwOut{Estimated brick center $c$, orientation $\theta$}
    
    \If{both $T_1$ and $T_2$ available} {
        Extract marker positions ${p}_1, {p}_2$ from $T_1, T_2$\;
        Refine $(\hat{p}_1, \hat{p}_2)$ using optimal filter\;
        $c \gets (\hat{p}_1 + \hat{p}_2)/2$\;
        $\theta \gets \arctan2(\hat{p}_{1,y} - \hat{p}_{2,y}, \; \hat{p}_{1,x} - \hat{p}_{2,x})$\;
    }
    \ElseIf{only $T_1$ available} {
        Extract position $p_1$, orientation $\phi_1$\;
        $c \gets p_1 + R(\phi_1) \cdot \text{offset}_1$\;
        $\theta \gets \phi_1 - \Delta$\;
    }
    \ElseIf{only $T_2$ available} {
        Extract position $p_2$, orientation $\phi_2$\;
        $c \gets p_2 + R(\phi_2) \cdot \text{offset}_2$\;
        $\theta \gets \phi_2 - \Delta$\;
    }
    \Else {
        No pose available, return None\;
    }
    
    \Return $(c, \theta)$\;
\end{algorithm}

Aiming further to eliminate the noise from the brick pose estimate and reduce the estimation error, the estimate is passed through a moving window filter, where it is smoothed out with a window equal to $N^W$ samples. 
To quantify the reliability of the position estimates, a confidence metric is also calculated based on the variability of recent measurements within the moving window of the most recent $N^W$ position estimates ${p_i = [x_i, y_i, z_i]}_{i=1}^{N^W}$. For each axis, the sample standard deviations $\sigma_x, \sigma_y, \sigma_z$ are computed. Since planar deviations are most critical for navigation, the maximum deviation in the horizontal plane is selected as $\sigma_{\max} = \max(\sigma_x, \sigma_y)$.
This deviation is then normalized with respect to a predefined tolerance $\sigma_{\text{tol}}$ and a confidence value $C \in [0,1]$ is defined~as:
\begin{equation}
C = 1 - \min\left(1, \frac{\sigma_{\max}}{\sigma_{\text{tol}}}\right),
\end{equation}

where $C=1$ corresponds to high certainty (low variability), and $C=0$ corresponds to low certainty (variability exceeding the tolerance).

\subsection{Descent Trajectory Planning for Brick Pickup and Place}
Once the pose of the brick has been estimated and the UAV is aligned with respect to it, the UAV starts a smooth and precise descent maneuver from its current hovering position to the center of the targeted brick.
Since terminal constraints are particularly stringent in pickup-and-place scenarios, a jerk-minimized trajectory \cite{kumar2024minimum} is considered to ensure smooth motion, with terminal accuracy enforced as hard constraints on the UAV during descent. This trajectory is obtained as the solution of the following optimal control problem: 
\begin{align}\label{costJ}
\min_{\mathbf{u}} \; J = \frac{1}{2} \int_{0}^{t_f} \mathbf{u}^T \mathbf{u} \, dt
\end{align}
subjected to, the UAV dynamics represented in the target fixed inertial frame of reference: $\dot{\mathbf{r}} = \mathbf{v}, \dot{\mathbf{v}} = \mathbf{a}, \dot{\mathbf{a}} = \mathbf{u}$
%
while enforcing the boundary constraints on position, velocity, and acceleration at both initial $(\mathbf{r}_0, \mathbf{v}_0, \mathbf{a}_0)$ and the terminal $(\mathbf{r}_f, \mathbf{v}_f, \mathbf{a}_f)$ states.
Here, $\mathbf{r} \in R^3$ denotes the position, $\mathbf{v} \in R^3$ indicates the velocity and $\mathbf{a} \in R^3$ represents the acceleration of the UAV in the target fixed frame. The jerk $\mathbf{u} \in R^3$, is introduced as a pseudo control variable in order to impose the terminal acceleration as a hard constraint in the trajectory design. The solution yields a closed-form trajectory, expressed as a quintic polynomial 
:
\begin{equation} \label{eq:UAV_Trajectory}
 {{\mathbf{r}}}({{t}_{go}})={{\mathbf{c}}_{{3}}}t_{go}^{5}+{{\mathbf{c}}_{{2}}}t_{go}^{4}+{{\mathbf{c}}_{{1}}}t_{go}^{3}+\mathbf{a}_f\frac{t_{go}^{2}}{2}-{\mathbf{v}_f}{{t}_{go}}+\mathbf{r}_f
 \end{equation}
 where, coefficients are expressed in closed form as:
\begin{align}\nonumber
\mathbf{c}_1 &= \frac{10}{t_{\text{go}}^3} Z_{r0} + \frac{4}{t_{\text{go}}^2} Z_{v0} + \frac{1}{2t_{\text{go}}} Z_{a0} \\
\mathbf{c}_2 &= -\frac{15}{t_{\text{go}}^4} Z_{r0} - \frac{7}{t_{\text{go}}^3} Z_{v0} + \frac{1}{t_{\text{go}}^2} Z_{a0} \\\nonumber
\mathbf{c}_3 &= \frac{6}{t_{\text{go}}^5} Z_{r0} + \frac{3}{t_{\text{go}}^4} Z_{v0} + \frac{1}{2t_{\text{go}}^3} Z_{a0},
\end{align}
\begin{align}\nonumber
Z_{r0} &= \mathbf{r}_0 - \mathbf{r}_f + \mathbf{v}_f t_{\text{go}} - \frac{1}{2} \mathbf{a}_f t_{\text{go}}^2, \\
Z_{v0} &= \mathbf{v}_0 - \mathbf{v}_f + \mathbf{a}_f t_{\text{go}}, \\\nonumber
Z_{a0} &= \mathbf{a}_0 - \mathbf{a}_f.
\end{align}
The parameter $t_{\text{go}}=(t_f-t)$ represents the remaining flight time. The trajectory in \eqref{eq:UAV_Trajectory} is used as a reference for the path-tracking controller, guiding the UAV from its current position to the target brick’s center while enforcing zero terminal velocity and the desired resultant acceleration over the finite horizon. This formulation maintains trajectory continuity up to the second derivative, allowing a smooth deceleration as the UAV approaches the target. Moreover, it enforces high terminal accuracy, facilitating robust grasping and pick-up. 
%

\subsection{UAV Control}
The motion of the UAV is governed by a hierarchical control architecture composed of two layers. At the low level, the onboard flight controller executes its internal velocity controller, which directly interfaces with the actuators to ensure stable flight. On top of this, a custom-designed PID-based motion controller is implemented to regulate the UAV’s position and orientation with respect to the mission objectives. The PID controller computes the desired velocity references based on the position error and dynamically compensates for disturbances, while the Pixhawk ensures the accurate tracking of these commands. This layered approach allows for both robustness and adaptability: the low-level controller guarantees stability and safety, whereas the high-level PID enhances tracking performance and enables seamless integration with the task planner. The overall control scheme has been tuned to balance responsiveness and smoothness, ensuring reliable operation under the dynamic conditions of aerial construction.

\section{Adhesion UAV}

\subsection{Platform Description}
\begin{figure}
    \centering
    \includegraphics[width=0.8\linewidth]{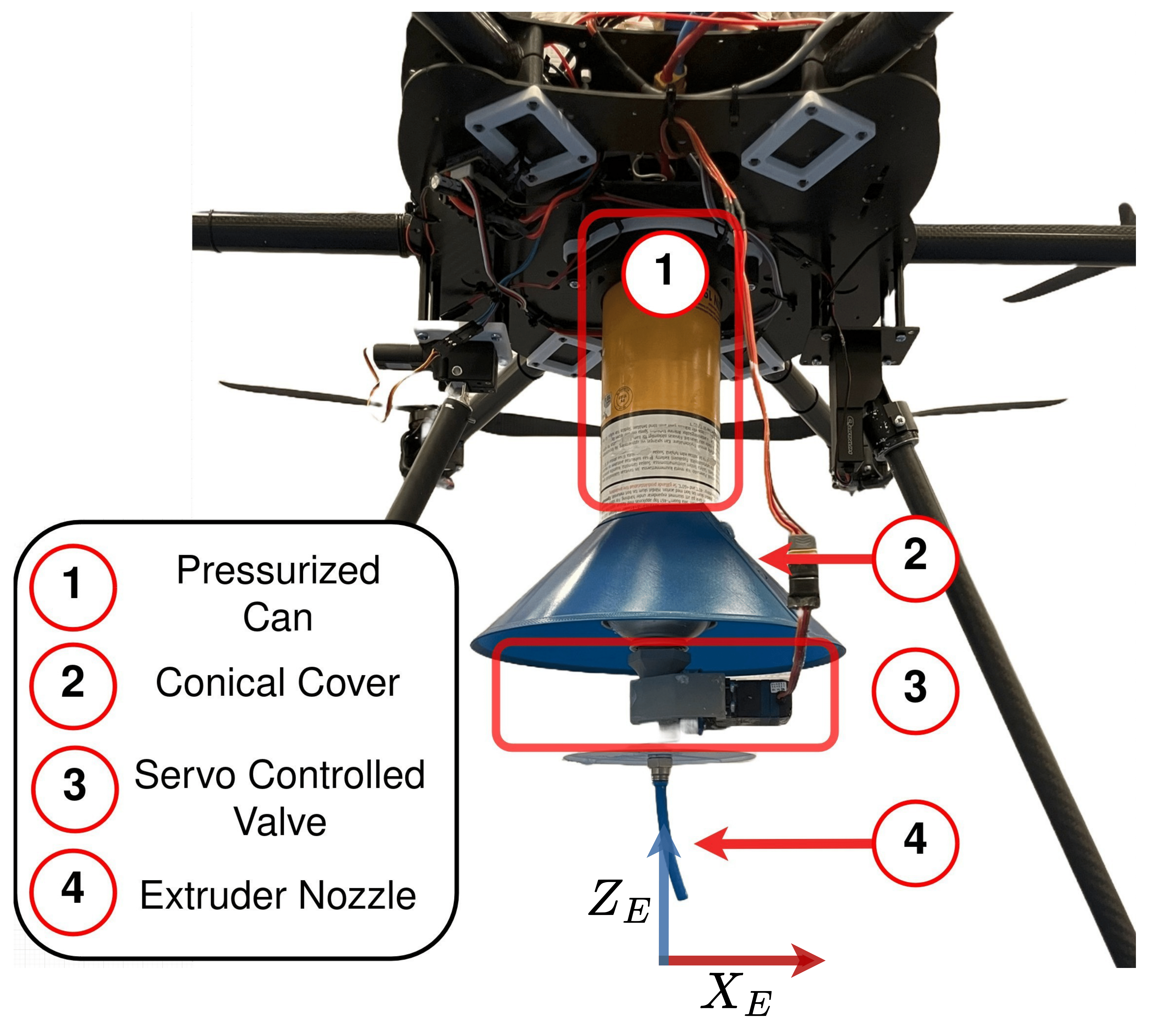}
    \caption{Adhesion UAV Spraying Mechanism}
    \label{fig:adhesionUAV}
    \vspace{-8mm}
\end{figure}

The material used for providing adhesion between the placed bricks is a single-compound expandable foam, which was selected due to its lightweight nature and adhesion properties. It is already placed in a commercial canister, ready to be sprayed.
To enable the spraying of the adhesion material, a modular 3D-printed spraying mechanism is developed and integrated with the aerial platform. The mechanism encapsulates the canister at both the top and bottom. The canister’s upper section is connected to a brass ball valve that regulates the outflow of material propelled by internal gases. As shown in Fig. \ref{fig:adhesionUAV}, the valve lever is mechanically coupled to a servo motor, secured by a 3D-printed bracket at the base of the canister. Controlled in real time by the onboard computer, the servo governs material flow, allowing precise activation and deactivation of extrusion during flight. A small plastic tube, attached to the valve outlet, completes the deposition system.
The extrusion mechanism is mounted beneath an off-the-shelf hexacopter platform with a wingspan of 0.76 m and a mass of 4.5 kg (including batteries). To avoid collisions with previously placed bricks and to also allow the extruder to operate as close as possible to the printing surface, retractable landing gear is employed and retracted during flight. Furthermore, to mitigate the adverse influence of propeller-induced downwash on material deposition, a protective cover and disk are placed around the valve to obstruct airflow. Although this measure does not entirely eliminate flow disturbances, it substantially reduces their impact on spraying quality.

\subsection{Adhesion Execution}
Once the adhesion UAV has been assigned an adhesion task $\mathcal{T}^A_j$, it takes off and elevates to cruising altitude, moves laterally, and hovers over the start of the adhesion position $P^A_j$ until it gets stable. Then a controlled descent starts in order to reach above the adhesion point with a vertical offset to account for the spraying action. It then proceeds with actuating the valve, starting the spraying of the adhesion material and moving smoothly along the length $l^A_j$ of the commanded adhesion. After it is completed, it continues to the next assigned task or goes back to its home position.

\vspace{-2mm}
\section{Results}
\begin{figure}[b]
    \centering
    \includegraphics[width=0.95\linewidth]{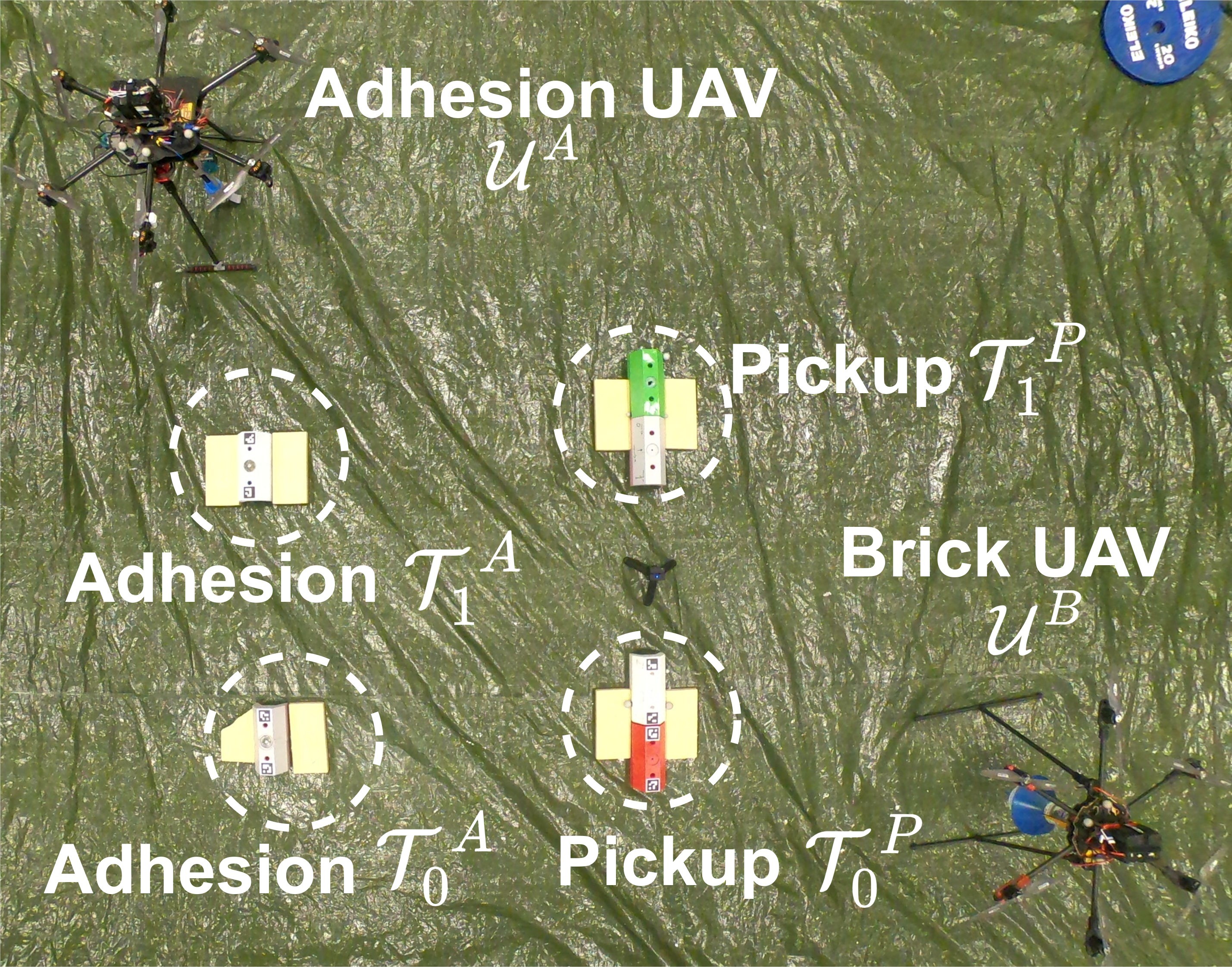}
    \caption{Top view of mission Workspace. Pickup points $\mathcal{T}^P_1, \mathcal{T}^P_2$ are located on the right, while the designated adhesion points $\mathcal{T}^A_0, \mathcal{T}^A_1$ are on the left.}
    \label{fig:workspace}
    \vspace{-5mm}
\end{figure}

\subsection{Experimental Setup}
The proposed framework is evaluated in a flight arena located in an indoor lab environment. Each UAV is equipped with a LattePanda 3 Delta, where the module responsible for the execution, movement, and control of the UAV were executed onboard. Additionally, each UAV is equipped with an onboard Pixhawk-Cube Orange low-level flight controller. The localization of the UAVs is provided by the Vicon motion tracking system and is fed wirelessly to the UAVs. Additionally, the wall construction is calculated offline before the start of the mission using the position feedback of the UAVs.
All the interfaces with the actuators of the robots are carried out using the onboard Arduino of Latte Panda.

\begin{figure}
    \centering
\includegraphics[width=0.75\linewidth]{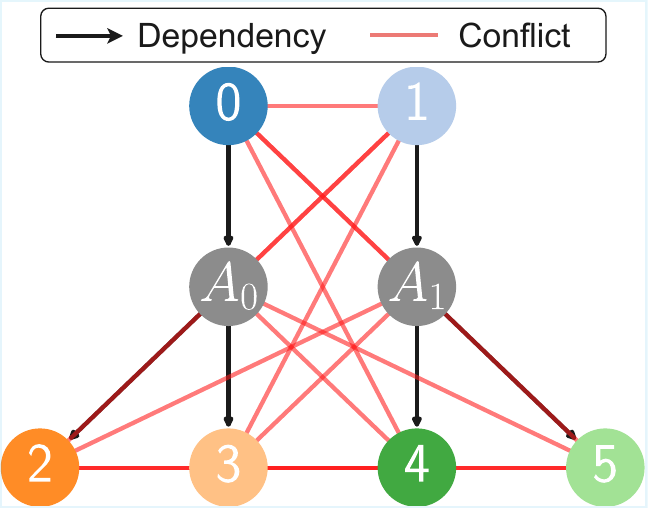}
    \caption{Dependency $\mathcal{G}^d$ and Conflict $\mathcal{G}^c$ Graph for the given mission.}
    \label{fig:depGraph}
    \vspace{-8mm}
\end{figure}

\begin{figure*}[b]
    \centering
    \includegraphics[width=0.9\linewidth]{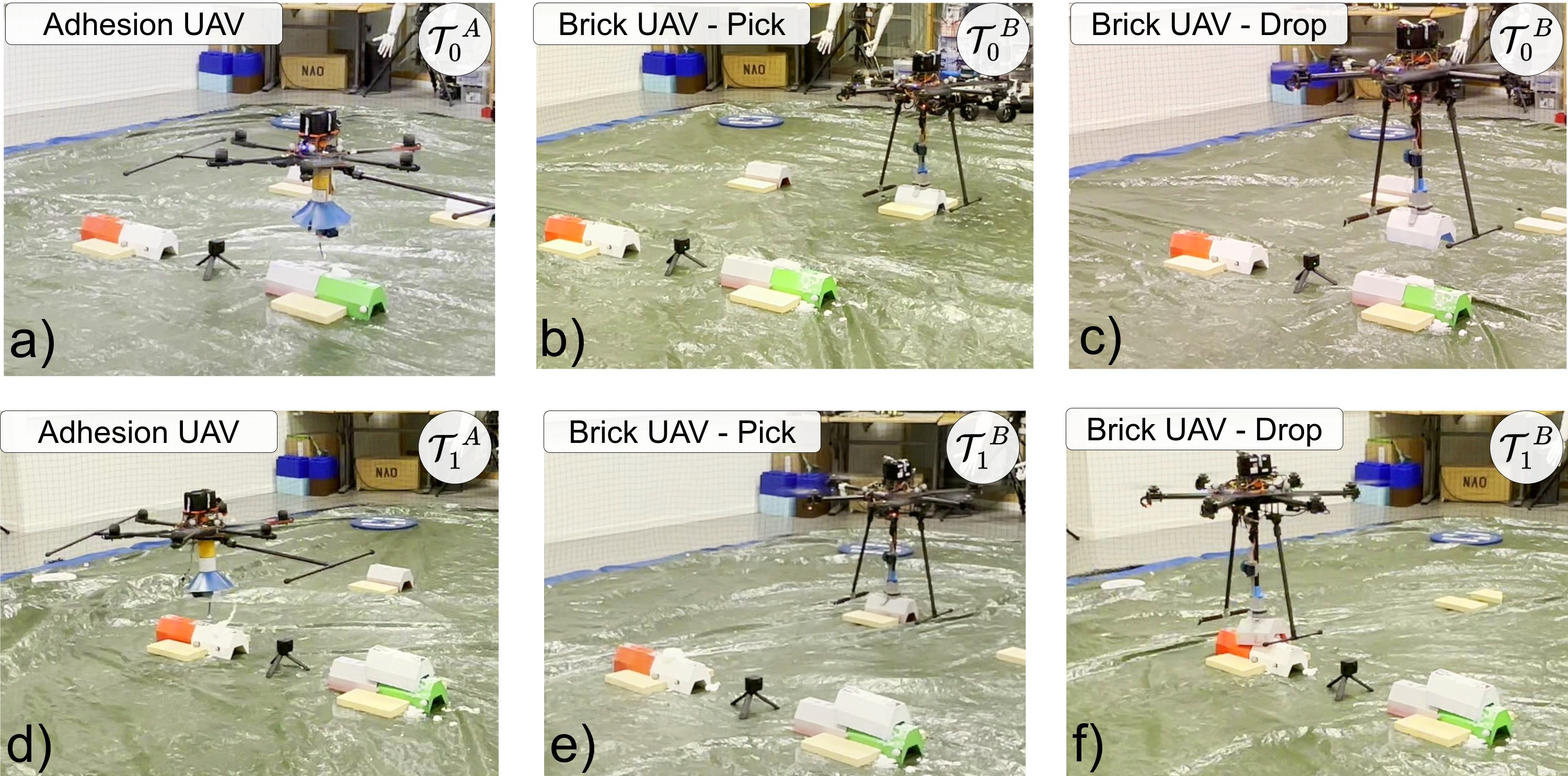}
    \caption{Sequential snapshots of the mission. Application of Adhesion to first drop off place by adhesion UAV (a), pickup of Brick 0 (b) and drop off at the designated point (c) while the adhesion on the second point (d), along with the pickup (e) and drop-off (f) of the second brick.} 
    \label{fig:missionSequence}
\end{figure*}

The distance between the two Aruco markers used for the deployment of the optimal filter is defined as $d=17.5$ cm, while the gains used for the deployment of the optimal filter are as follows $w_{\text{prox}} = 2$, $w_{\text{dist}} = 1$ and $w_{\text{plane}} = 1$. 
The sampling window for the moving average filter is selected to be $N^w=30$ while the maximum deviation is set to $\sigma_{tol}=$ \SI{4}{cm} and the confidence metric threshold set to $C_{th} = 0.75$.
The minimum clearance between the UAVs for the deployment is selected to be equal to $r_c= 1.5 m$ given that the wingspan of both of them is approximately $0.8$ m. The cruising altitude of the UAV is $h_{cr} = 1.2$ m. while the duration for the smooth guided descent is set to $t_f=7$ s.

\subsection{Brick Estimation Evaluation}
To assess the accuracy of the proposed brick pose estimation algorithm, the estimated brick position was compared against ground-truth data obtained from the motion capture system. Across multiple trials, the average error was found to be approximately \SI{2}{cm} per axis, corresponding to a mean 2D error of \SI{3}{cm}. This level of accuracy lies within the margins required for reliable brick pickup and placement, as the end-effector mechanism is specifically designed to tolerate minor position and orientation deviations by passively aligning 
the pickup point with the target brick upon contact.

\subsection{Case Study Mission}

\begin{figure*}[h]
    \centering
    \begin{subfigure}{0.28\textwidth}
        \includegraphics[width=\linewidth]{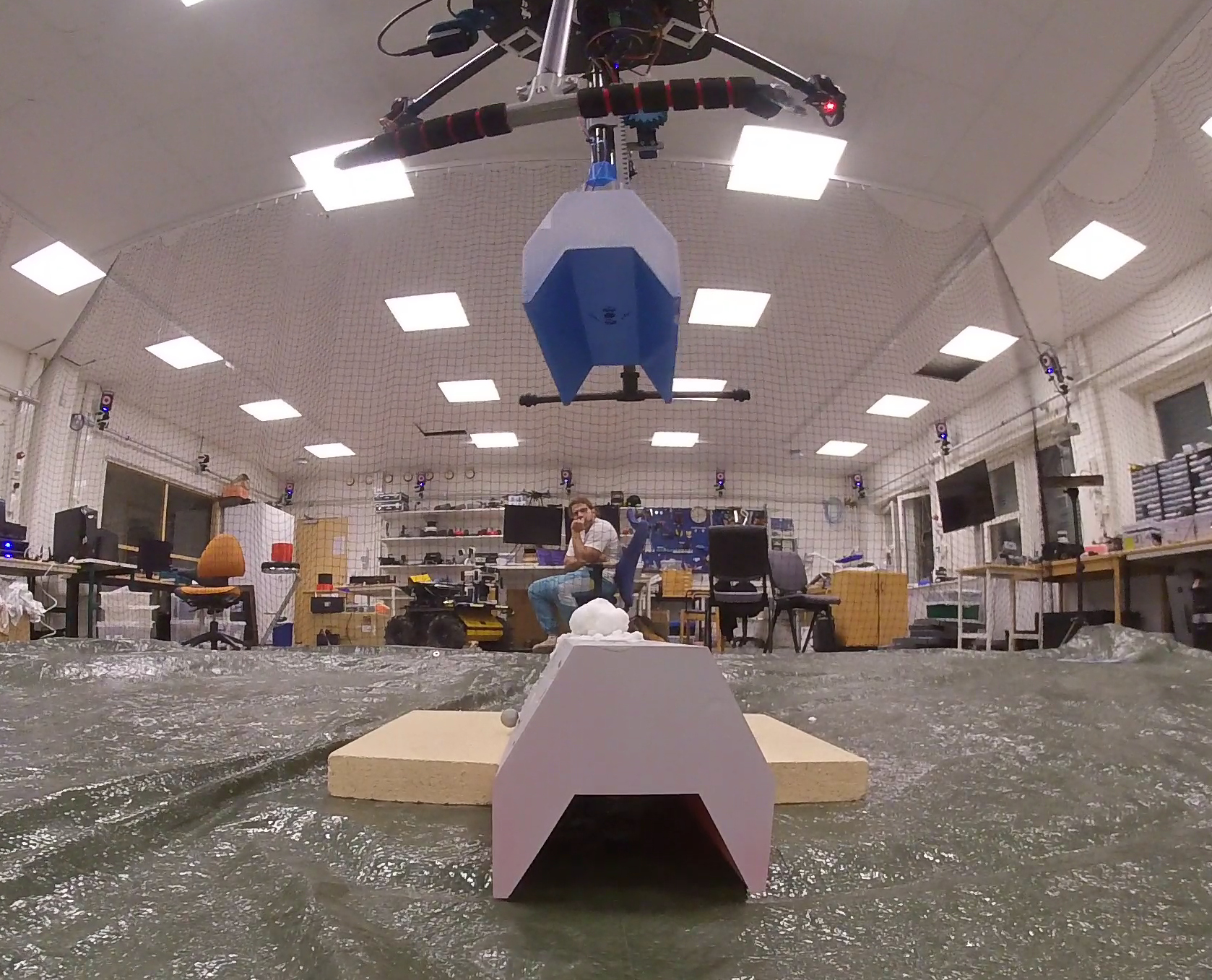}
        \caption{Hover Above}
        \label{fig:first}
    \end{subfigure}
    \hfill
    \begin{subfigure}{0.28\textwidth}
        \includegraphics[width=\linewidth]{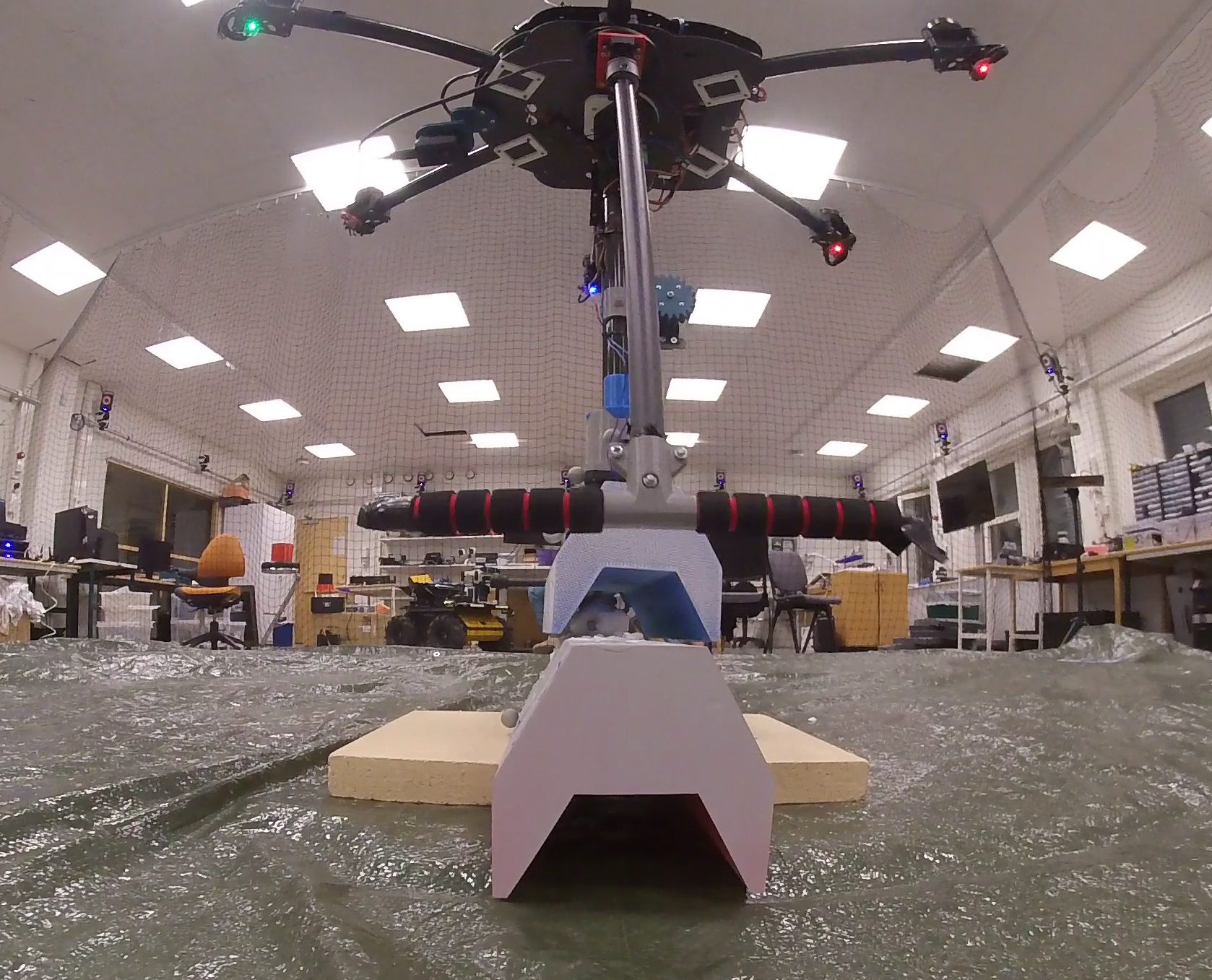}
        \caption{Smooth Descent}
        \label{fig:second}
    \end{subfigure}
    \hfill
    \begin{subfigure}{0.28\textwidth}
        \includegraphics[width=\linewidth]{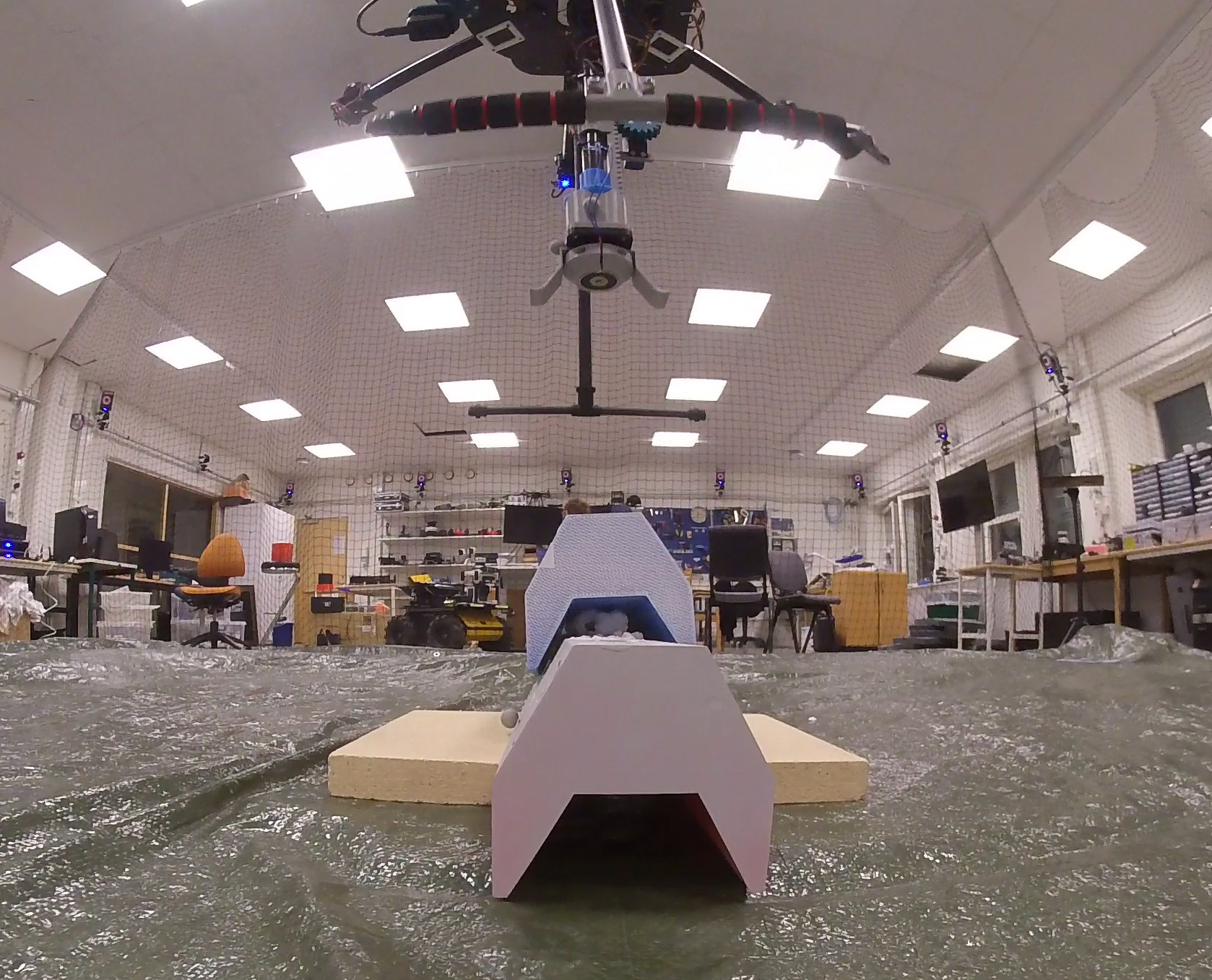}
        \caption{Drop-off}
        \label{fig:third}
    \end{subfigure}
    \caption{Side view of the drop-off sequence.}    \label{fig:dropOffSeq}
    \vspace{-8mm}
\end{figure*}

Aiming to evaluate the execution of the proposed framework, a case study mission is carried out. Specifically, as shown in Fig. \ref{fig:workspace}, an adhesion UAV $U^A$ along with a pickup UAV $U^P$ are deployed. There are two adhesion tasks $\mathcal{T}^A_0, \mathcal{T}^A_1$ that also coincide with two drop-off tasks $\mathcal{T}^D_0,\mathcal{T}^D_1$ on top of two already placed pairs of bricks on the left. While the approximate locations of the two pick-up tasks $\mathcal{T}^P_0,\mathcal{T}^P_1$ are on the right. The calculated dependency graph $\mathcal{G}^d$ for the given construction blueprint is shown in Fig. \ref{fig:depGraph}.

Initially, only the adhesion tasks $\mathcal{T}^A_0$ and $\mathcal{T}^A_1$ are available. Consequently, the reactive mission planner assigns task $\mathcal{T}^A_0$ to the adhesion UAV, which takes off and moves all the way till the start of the the adhesion placement and then deposits the adhesive material onto the designated area between the bricks, as illustrated in Fig.~\ref{fig:missionSequence}.a. Upon completion of this task, UAV $\mathcal{U}^A$ notifies the mission planner that its assigned task is completed successfully and the dependency graph $\mathcal{G}^d$ is dynamically updated. Thus, the brick task $\mathcal{T}^B_0$ becomes independent of the adhesion task and is therefore reactively assigned to the brick UAV $\mathcal{U}^B$ while at the same time the conflict graph $\mathcal{G}^c$ is also updated, rendering all the tasks in conflict with $\mathcal{T}^B_0$ unavailable. Although task $\mathcal{T}^A_1$ is also eligible for assignment from a dependency standpoint, it cannot be executed simultaneously with the brick task because their proximity violates the minimum clearance constraint $r_c$. UAV $\mathcal{U}^B$ collects the brick from the designated pickup point (see Fig.~\ref{fig:missionSequence}.b) and proceeds toward the drop-off location. The subsequent drop-off sequence is illustrated in Fig.~\ref{fig:dropOffSeq}, where the UAV first hovers above the target, then smoothly descends, and finally releases the brick once the placement error $e_{pl}$ falls below the required threshold. Since no further tasks are available at that point, the UAV returns to its home position. An analogous procedure is followed for the second adhesion and drop-off tasks. Sequential snapshots of the entire execution are presented in Fig.~\ref{fig:missionSequence}.
The attached video demonstrates the execution of the discussed construction mission in a lab environment.

\section{Conclusion}
This work presents a fully autonomous aerial masonry construction framework using heterogeneous UAVs. A pickup-drop UAV for brick manipulation and an adhesion UAV for adhesion application were developed. These UAVs collaborate through a reactive planning framework that integrates a dependency graph, conflict graph, and hierarchical state machines to enable robust and coordinated execution. Minimum-jerk trajectories and real-time brick pose estimation using ArUco markers along with a least-squares filter ensure precise pickup and smooth motion. Experimental results validate the feasibility of the proposed system, demonstrating efficient and robust autonomous masonry operations, laying the foundation for future developments in aerial robotic construction.
To the best of the authors’ knowledge, this is the first experimental demonstration of autonomous aerial masonry with heterogeneous UAVs, featuring one for brick placement and another for adhesion application.
\vspace{-3mm}
\FloatBarrier
\bibliographystyle{IEEEtran}
\bibliography{sample}

\begin{thebibliography}{10}
\providecommand{\url}[1]{#1}
\csname url@samestyle\endcsname
\providecommand{\newblock}{\relax}
\providecommand{\bibinfo}[2]{#2}
\providecommand{\BIBentrySTDinterwordspacing}{\spaceskip=0pt\relax}
\providecommand{\BIBentryALTinterwordstretchfactor}{4}
\providecommand{\BIBentryALTinterwordspacing}{\spaceskip=\fontdimen2\font plus
\BIBentryALTinterwordstretchfactor\fontdimen3\font minus \fontdimen4\font\relax}
\providecommand{\BIBforeignlanguage}[2]{{%
\expandafter\ifx\csname l@#1\endcsname\relax
\typeout{** WARNING: IEEEtran.bst: No hyphenation pattern has been}%
\typeout{** loaded for the language `#1'. Using the pattern for}%
\typeout{** the default language instead.}%
\else
\language=\csname l@#1\endcsname
\fi
#2}}
\providecommand{\BIBdecl}{\relax}
\BIBdecl

\bibitem{advancementsRoboticsConstr}
B.~Xiao, C.~Chen, and X.~Yin, ``Recent advancements of robotics in construction,'' \emph{Automation in Construction}, vol. 144, p. 104591, 2022.

\bibitem{saidi2016}
K.~Saidi, T.~Bock, and C.~Georgoulas, ``Robotics in construction,'' in \emph{Springer Handbook of Robotics}.\hskip 1em plus 0.5em minus 0.4em\relax Cham: Springer International Publishing, 2016, pp. 1493--1520.

\bibitem{largeScale3DPrint}
V.~Mechtcherine, V.~N. Nerella, F.~Will, M.~Näther, J.~Otto, and M.~Krause, ``Large-scale digital concrete construction – conprint3d concept for on-site, monolithic 3d-printing,'' \emph{Automation in Construction}, vol. 107, p. 102933, 2019.

\bibitem{BRIX}
P.~Ruttico, M.~Pacini, and C.~Beltracchi, ``Brix: an autonomous system for brick wall construction,'' \emph{Construction Robotics}, vol.~8, no.~10, 2024.

\bibitem{brickLabyrinth}
L.~Pi{\v{s}}korec, D.~Jenny, S.~Parascho, H.~Mayer, F.~Gramazio, and M.~Kohler, ``The brick labyrinth,'' in \emph{Robotic Fabrication in Architecture, Art and Design 2018}, J.~Willmann, P.~Block, M.~Hutter, K.~Byrne, and T.~Schork, Eds., 2019, pp. 489--500.

\bibitem{printingWhileMoving}
M.~E. Tiryaki, X.~Zhang, and Q.-C. Pham, ``Printing-while-moving: a new paradigm for large-scale robotic 3d printing,'' in \emph{2019 IEEE/RSJ International Conference on Intelligent Robots and Systems (IROS)}, 2019, pp. 2286--2291.

\bibitem{nature_aerial_AM}
K.~Zhang, P.~Chermprayong, F.~Xiao, and D.~e.~a. Tzoumanikas, ``{Aerial additive manufacturing with multiple autonomous robots},'' \emph{Nature}, vol. 609, no. 7928, pp. 709--717, 2022.

\bibitem{stamatopoulos2025ExperimentAutCon}
M.-N. Stamatopoulos, J.~Haluška, E.~Small, J.~Marroush, A.~Banerjee, and G.~Nikolakopoulos, ``Fully autonomous chunk-based aerial additive manufacturing with offset-free predictive control,'' \emph{Automation in Construction}, vol. 178, p. 106361, 2025.

\bibitem{stamatopoulos2024collaborative}
M.-N. Stamatopoulos, A.~Banerjee, and G.~Nikolakopoulos, ``{Collaborative Aerial 3D Printing: Leveraging UAV Flexibility and Mesh Decomposition for Aerial Swarm-Based Construction},'' in \emph{2024 International Conference on Unmanned Aircraft Systems (ICUAS)}.\hskip 1em plus 0.5em minus 0.4em\relax IEEE, 2024, pp. 45--52.

\bibitem{flightAssembledArch}
F.~Augugliaro, S.~Lupashin, M.~Hamer, C.~Male, M.~Hehn, M.~W. Mueller, J.~S. Willmann, F.~Gramazio, M.~Kohler, and R.~D'Andrea, ``{The flight assembled architecture installation: Cooperative contruction with flying machines},'' \emph{IEEE Control Systems}, vol.~34, no.~4, pp. 46--64, 2014.

\bibitem{Xie2025aerial}
X.~Xie, S.~Gu, Y.~Xu, and P.~F. Yuan, ``Aerial arch bridge construction: Precise and flexible assembly with drones using a 6dof parallel manipulator,'' in \emph{The 43rd Education and Research in Computer Aided Architectural Design in Europe Conference, eCAADe 2025}, 2025.

\bibitem{feasibility}
S.~Goessens, C.~Mueller, and P.~Latteur, ``Feasibility study for drone-based masonry construction of real-scale structures,'' \emph{Automation in Construction}, vol.~94, pp. 458--480, 2018.

\bibitem{mbzircWallSaska}
T.~Baca, R.~Penicka, P.~Stepan, M.~Petrlik, V.~Spurny, D.~Hert, and M.~Saska, ``Autonomous cooperative wall building by a team of unmanned aerial vehicles in the mbzirc 2020 competition,'' \emph{Robotics and Autonomous Systems}, vol. 167, p. 104482, 2023.

\bibitem{mbzircMetaheuristicWallplanner}
B.~Elkhapery, R.~Pěnička, M.~Němec, and M.~Siddiqui, ``Metaheuristic planner for cooperative multi-agent wall construction with uavs,'' \emph{Automation in Construction}, vol. 152, p. 104908, 2023.

\bibitem{mbzircBarbara}
M.~Krizmancic, B.~Arbanas, T.~Petrovic, F.~Petric, and S.~Bogdan, ``Cooperative aerial-ground multi-robot system for automated construction tasks,'' \emph{IEEE Robotics and Automation Letters}, vol.~5, no.~2, pp. 798--805, 2020.

\bibitem{Xie2025Brickpilot}
X.~Xie, S.~Gu, Y.~Xu, and P.~F. Yuan, ``Brickpilot: Autonomous aerial bricklaying using drones and computer vision,'' in \emph{The 30th International Conference of the Association for Computer‑Aided Architectural Design Research in Asia, CAADRIA 2025}, vol.~2, 2025, pp. 203--212.

\bibitem{elkhapery2022exploratory}
B.~Elkhapery and E.~Small, ``Exploratory research towards automated masonry construction using uavs,'' in \emph{IOP Conference Series: Materials Science and Engineering}, vol. 1218, no.~1.\hskip 1em plus 0.5em minus 0.4em\relax IOP Publishing, 2022, p. 012005.

\bibitem{stamatopoulos2025safety}
M.-N. Stamatopoulos, S.~Velhal, A.~Banerjee, and G.~Nikolakopoulos, ``Safety-aware optimal scheduling for autonomous masonry construction using collaborative heterogeneous aerial robots,'' \emph{arXiv preprint arXiv:2506.18697}, 2025.

\bibitem{aruco}
S.~Garrido-Jurado, R.~Muñoz-Salinas, F.~Madrid-Cuevas, and M.~Marín-Jiménez, ``Automatic generation and detection of highly reliable fiducial markers under occlusion,'' \emph{Pattern Recognition}, vol.~47, no.~6, pp. 2280--2292, 2014.

\bibitem{kumar2024minimum}
P.~S. Kumar and R.~Padhi, ``Minimum jerk guidance for autonomous soft-landing of an unmanned aerial vehicle on a moving platform,'' in \emph{2024 18th International Conference on Control, Automation, Robotics and Vision (ICARCV)}.\hskip 1em plus 0.5em minus 0.4em\relax IEEE, 2024, pp. 1201--1206.

\end{thebibliography}

\end{document}